\documentclass[journal]{IEEEtran}
\ifCLASSINFOpdf
\else
\fi
\usepackage{url}
\usepackage{cite}
\usepackage{amsmath,amssymb,amsfonts}
\usepackage{mathtools}
\usepackage{bbm}
\usepackage{graphicx}
\usepackage{textcomp}
\usepackage{multicol}
\usepackage{multirow}
\usepackage{algorithm}
\usepackage[noend]{algpseudocode}
\usepackage{tabularx}
\usepackage{adjustbox}  
\usepackage{dblfloatfix}
\usepackage[para]{threeparttable}
\usepackage{caption}
\usepackage{stackengine}
\usepackage{xcolor}

\usepackage{subfig}
\usepackage{graphicx}
\usepackage{fancyhdr}
\usepackage{mathtools}

\usepackage[final]{changes}
\usepackage{changes}
\usepackage{lipsum}
\definechangesauthor[name={Nair}, color=orange]{Nair}

\def\x{{\mathbf x}}

\def\x{{\bf x}}

\def\K{{\bf K}}

\def\x{{\mathbf x}}

\def\w{{\bf w}}

\fancypagestyle{github}{\fancyhf{}\fancyfoot[L]{$^*$https://github.com/aimlab-wustl/GiniSVMMicro}}

\begin{document}
%
\title{Multiplierless In-filter Computing for tinyML Platforms}
%
%
%

\author{Abhishek~Ramdas~Nair$^*$,~\IEEEmembership{Member,~IEEE,}
        Pallab~Kumar~Nath$^*$,~\IEEEmembership{Member,~IEEE,}
        Shantanu~Chakrabartty,~\IEEEmembership{Senior Member,~IEEE,}
        and~Chetan~Singh~Thakur,~\IEEEmembership{Senior Member,~IEEE}
\thanks{Abhishek Ramdas Nair and Chetan Singh Thakur are with the Department
of Electronic Systems Engineering, Indian Institute of Science, Bangalore,
KA, 560012 INDIA e-mail: (abhisheknair@iisc.ac.in, csthakur@iisc.ac.in).}
\thanks{Pallab Kumar Nath was with Department of Electronic Systems Engineering, Indian Institute of Science and currently associated with Department of Electronics Communication Engineering, Pandit Deendayal Energy University,  Gandhinagar email: (pallab.nath@sot.pdpu.ac.in)} 
\thanks{Shantanu Chakrabartty is with Department of Electrical and Systems Engineering, Washington University in St. Louis, USA, 63130 e-mail: (shantanu@wustl.edu).}
\thanks{$^*$ Both Abhishek Ramdas Nair and Pallab Kumar Nath have contributed equally to this paper.}%
\thanks}

%
%

\markboth{}%
{A.R.Nair \MakeLowercase{\textit{et al.}}: Multiplierless In-filter Computing for tinyML Platforms}
%



\maketitle

\begin{abstract}
Wildlife conservation using continuous monitoring of environmental factors and biomedical classification, which generate a vast amount of sensor data, is a challenge due to limited bandwidth in the case of remote monitoring. It becomes critical to have classification where data is generated, and only classified data is used for monitoring. We present a novel multiplierless framework for in-filter acoustic classification using Margin Propagation (MP) approximation used in low-power edge devices deployable in remote areas with limited connectivity. The entire design of this classification framework is based on template-based kernel machine, which include feature extraction and inference, and uses basic primitives like addition/subtraction, shift, and comparator operations, for hardware implementation. Unlike full precision training methods for traditional classification, we use MP-based approximation for training, including backpropagation mitigating approximation errors. The proposed framework is general enough for acoustic classification. However, we demonstrate the hardware friendliness of this framework by implementing a parallel Finite Impulse Response (FIR) filter bank in a kernel machine classifier optimized for a Field  Programmable  Gate  Array (FPGA). The FIR filter acts as the feature extractor and non-linear kernel for the kernel machine implemented using MP approximation and a downsampling method to reduce the order of the filters. The FPGA implementation on Spartan 7 shows that the MP-approximated in-filter kernel machine is more efficient than traditional classification frameworks with just less than 1K slices.
\end{abstract}

\begin{IEEEkeywords}
IoT, FPGA, Filtering, Edge Computing.
\end{IEEEkeywords}

%
\IEEEpeerreviewmaketitle

\section{Introduction}
%
%
%
%

\IEEEPARstart{O}{ne} of the biggest challenges in biomedical classification is capturing data from different biosensors and providing interpretable information to improve diagnosis \cite{tobore2019deep} \cite{zhao2020role}. On the other hand, in the case of wildlife conservation, identifying and localizing the threatened species and providing supportive or corrective measures to enable restoration is a challenge \cite{witmer2005wildlife}. Emerging technologies in edge computing devices like low-power wireless sensor networks are currently being used in agriculture \cite{zgank2021iot} and healthcare, \cite{quintana2017low} in combination with Machine Learning (ML) techniques, known as tinyML. Most of the edge-based sensor data are time-series, and it has been proven that such data can be efficiently used for tinyML classification \cite{NairTemplate}. This type of classification can be applied to healthcare with Electrocardiogram (ECG), Electroencephalogram (EEG), Electromyography (EMG), and other time-series biomedical sensor data \cite{zhao2020role}.

\begin{figure}[]
\centerline{\includegraphics[page=1,scale=0.24,trim=0 0 0 0,clip]{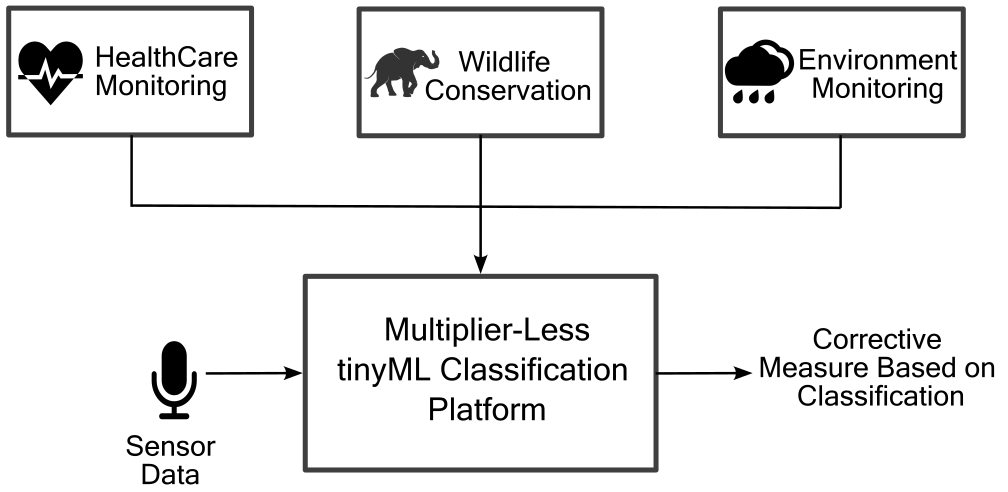}}
\caption{Ecological Conservation and Corrective System.}
\label{Fig:Intro}
\end{figure}

Similar systems have been in place for ecological \cite{popovic2017architecting} and marine monitoring \cite{luo2021localization}. Acoustic based classification using wireless biosensors have been proven highly efficient in case of ecological applications \cite{de2013wireless} \cite{tuia2022perspectives} as shown in Fig.\ref{Fig:Intro}. These sensors may generate a high amount of data, but the relevant training data will be sparse, like in the case of rare or near-extinct species detection \cite{shan2006machine}. Hence, classification at the sensor node becomes even more critical as large data transmission over the network will require higher bandwidth. Despite performing well for a high volume of data, Deep Neural Networks (DNNs) do not generalize well in IoT applications, as training data is rare \cite{heydari2018effect}. Moreover, training a DNN requires high-powered systems to generate appropriate learned parameters. IoT-based edge devices typically have limited resources and power. Hence, training on edge is highly improbable. However, IoT systems can have inference executed on devices using quantized and pruned versions of the neural networks  \cite{han2015deep}. Hence, such wireless sensors must perform classification at the edge and be tunable while having the lowest possible area and power. 

Machine learning techniques like Support Vector Machines (SVMs), K-Nearest Neighbour (KNN), and kernel machines have proven to be robust and interpretable for rare event classification \cite{tang2008svms} \cite{nair2019hybrid} \cite{maalouf2011robust}. However, these techniques have traditionally been computationally intensive for training and inference. As most of the computation is based on Matrix-Vector Multiplication (MVM) operation, replacing multipliers with more fundamental basic primitives like addition/subtraction will enable to design energy-efficient classification framework \cite{horowitz20141}. Multipliers tend to produce a precision explosion. We can exploit the computational primitives and approximations inherent in digital units like counters, underflow/overflow, and additions/subtraction by replacing them. In literature, there have been ways to tackle precision explosion due to multiplication for multiply-accumulate operations like quantization \cite{guo2018survey} or representation in a new number system \cite{tagliavini2018transprecision}.

Traditionally, IoT-based machine learning and neural networks train offline with full precision and deploy the inference at lower precision fixed point \cite{hubara2017quantized}. Even with quantization-aware training, the backpropagation in training is done in full precision, and only the forward pass is quantized \cite{fan2020training}. This may be an efficient training technique, but it still is expensive for re-training on the IoT platform. There have been instances where the gradients in backpropagation have been quantized \cite{rastegari2016xnor} \cite{zhou2016dorefa}. However, these systems fail to achieve convergence during backpropagation \cite{guo2018survey}, or they work well only when training data is available in large numbers. Moreover, the front-end, like filters and feature extractors in most edge devices, is implemented at higher precision with only the classifier quantized.

This paper leverages the energy-efficient bird density detection tinyML system \cite{hemanth2022} which uses the in-filter computing \cite{NairTemplate} with template-based SVM architecture \cite{kumar2019neuromorphic}. Here we apply the Margin Propagation (MP) principle \cite{chakrabartty2004margin} to this architecture to develop a multiplierless in-filter computing framework, which exploits the computing and nonlinear primitives in the feature extraction process. Our goal is to exploit the hardware efficiency due to MP approximation for designing multiplierless filtering and classification operations, including nonlinear transformations using the kernel functions, also used as a feature extractor for training and inference. The multiplierless MP-based kernel machine has been proven to provide energy-efficient classification \cite{nair2022multiplierless}. 
In our design, we implement an FIR filter bank, used as a feature extractor and kernel function, arranged in a multi-rate frequency model \cite{singh2018car} using a multiplierless approach based on MP. This model reduces the FIR filter order used in the filter bank and helps achieve a low computation footprint. We believe that our proposed framework has the following key advantages:
\begin{itemize}
  \item  End-to-end multiplierless framework for acoustic classification using only basic primitives like addition/subtraction, underflow/overflow, shift, and comparison operations.
  \item Feature extraction and kernel function are combined to form an efficient computational system.
  \item Scalable system with user-defined memory footprint based on IoT hardware constraints.
  \item Integrated training using MP-based approximation mitigates approximation errors introduced in filtering and classifier.
  \item \added{ Since our framework uses basic computational primitives (no multipliers), it enables the implementation to push for much higher clock frequency (in this case 166MHz).}
\end{itemize}
We have implemented the inference framework on an FPGA as proof of concept IoT implementation. We have validated our architecture on the environmental sound dataset \cite{piczak2015dataset}, which showcases the capabilities of potential deployment to identify wildlife sounds or even sounds that may indicate possible poaching or timber smuggling. 

The rest of this paper is organized as follows. Section II provides a brief discussion of related work, followed by section III, where we present the in-filter computation using an MP kernel machine. Section IV provides the FPGA implementation details.
Section V provides results with an audio-based dataset for detection and surveillance applications. Section VI concludes this paper, provides some useful applications and discusses possible future work using this framework.


\section{Related Works}
A framework to build automated animal recognition in the wild, aiming at an automated wildlife monitoring system, has been discussed in \cite{nguyen2017animal}. The authors propose to use camera traps to capture data which becomes increasingly difficult in low light conditions and requires regular maintenance with increased power consumption.
Extensive studies have been done using animal sounds for the recognition and classification of species employing SVMs and Hidden Markov Models (HMM) \cite{weninger2011audio}. However, hardware implementation of such systems is not energy efficient for wildlife deployment.
An efficient hardware implementation of acoustic classification for biometric security access was realized in \cite{ramos2009svm}, where Mel-Frequency Cepstral Coefficients (MFCC) is used as a feature extractor and implemented on-chip along with an SVM classifier. The MFCC and the SVM kernel occupy a high amount of hardware. \deleted{(5351 registers and 6781 Look-Up Tables (LUTs)). }
To improve the hardware efficiency, we can explore the method where we can combine the feature extraction and SVM kernel as a single function as described in \cite{NairTemplate}. This eliminates the use of separate feature extraction and improves the hardware efficiency in terms of power (8 mW) and area \deleted{(2864 registers and 1517 LUTs)}. The work in \cite{hemanth2022} shows efficient microcontroller implementation of the work described in \cite{NairTemplate} for classification in case of ecological applications.

We still have a scope to improve the system's efficiency by exploring approximate computing to replace traditional resource-heavy operations like multipliers with more basic operators like additions. In literature, we have seen many approximate techniques like Canonic Signed Digits (CSD) \cite{mandal2014implementation}, logarithmic function using look-up tables \cite{xue2020real} or powers of 2 approximation \cite{elhoushi2021deepshift}. These systems do not offer a complete end-to-end multiplierless computation. 

Traditionally, energy efficiency can be achieved using a quantization technique instead of approximate computation, like fixed-point precision \cite{lin2016fixed}. However, such linear quantization techniques result in accuracy loss as the data is represented uniformly. To mitigate this error, adaptive quantization can be used as described in \cite{zhou2018adaptive}. This technique uses a measurement model to estimate the correct quantization for all the parameters of the classification system. Implementation of such a system is not feasible in the case of edge devices as there is the overhead of estimating the quantization levels, compromising the device's energy efficiency. Representation of the entire framework in the bfloat number system can provide similar effects of adaptive quantization \cite{hagiescu2019bfloat}.  

\begin{figure}[]
\centerline{\includegraphics[page=1,scale=0.2,trim=0 0 0 0,clip]{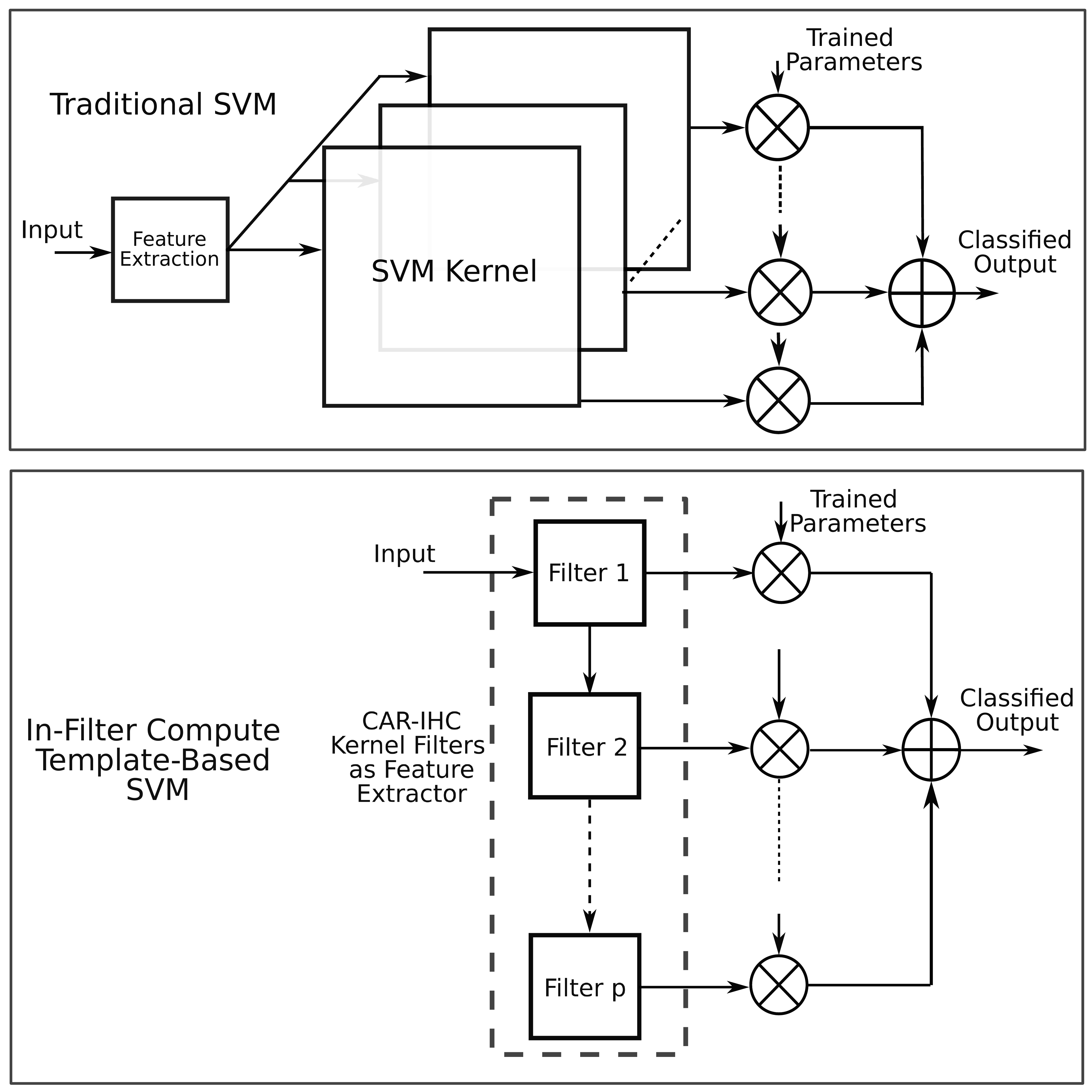}}
\caption{Comparison of Traditional SVM and In-Filter Compute Template based SVM. The Template based SVM, using $p$ filters, enables user-specified hardware constraints and uses feature extractor as the Kernel Function \cite{NairTemplate}.}
\label{Fig:T-SVM}
\end{figure}

\begin{figure*}[ht]
\centerline{\includegraphics[page=1,scale=0.45,trim=0 0 0 0,clip]{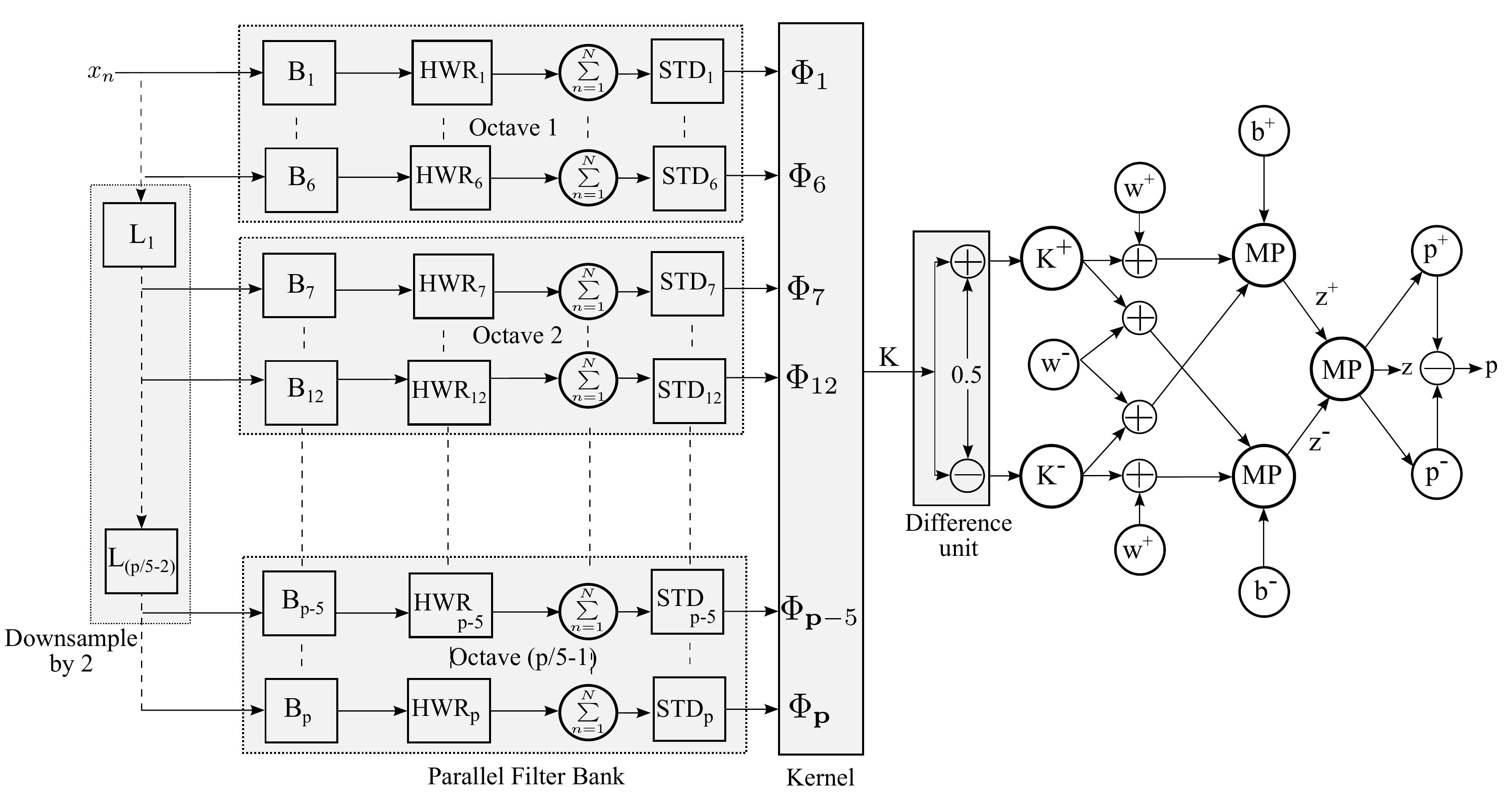}}
\caption{ FIR parallel filter bank framework for MP based classification for $p \times 1$ Kernel. \added{Here, the input $x_n$ is provided to the parallel FIR filter bank to generate a $p \times 1$ kernel. This kernel is used as input for a single layer classification network formed using MP modules. The parallel filter bank and the downsampling low pass filter blocks also use MP modules for computation.}}
\label{Fig:Logic_Block_Diag}
\end{figure*} 


In \cite{nair2022multiplierless}, we see an end-to-end multiplierless system using the MP approximation technique for kernel machines. We extend the capabilities of this framework in our current work where we use the feature extraction used as kernel function, as used in \cite{NairTemplate} and \cite{hemanth2022}, and implement this kernel in MP approximation using the MP principle in \cite{nair2022multiplierless}.  
The resulting framework is a one-of-a-kind digital hardware implementation of a multiplierless acoustic classifier with a feature extractor used as a kernel.

\section{In-Filter Computation using Margin Propagation Kernel Machine} \label{MPcomp}
In-filter computation described in \cite{NairTemplate} and \cite{hemanth2022}, combines the feature extraction and non-linear SVM kernel into a single function \cite{kumar2019neuromorphic} as opposed to a traditional SVM as shown in Fig.\ref{Fig:T-SVM}. We leverage this principle, use an FIR filter as the kernel function, and implement this framework using MP-based approximation.
MP-based kernel machine has proven to be an energy-efficient system for implementing a classification framework for edge devices \cite{nair2022multiplierless}.

\subsection{Precision Explosion and MP as Adaptive Quantization}
Consider a fully connected single hidden layer network. In the case of a fully connected network, for a 8-bit quantized inputs, the hidden layer output will generate 16-bit output from multiplication. For the 16-dimension input vector, the hidden layer will generate 20-bit output, and the final output layer will have 44-bit output. If we quantize this output to 8-bit, we may lose a substantial amount of accuracy.
In contrast, the MP operation, which involves addition, would generate 18-bit at the output, which would result in lower accuracy loss. Thus, MP avoids numerical precision explosion caused by conventional MAC operations. Moreover, the quantization error can be mitigated by having an online training system, which has been shown in \cite{nair2022multiplierless}.

\captionsetup[subfigure]{labelformat=empty}
\begin{figure*}[ht]
    \centering
     \subfloat{\label{Fig:NoDown}\stackinset{l}{.01in}{t}{.07in}{(a)}{\includegraphics[page=1,scale=0.15,trim=0 0 0 0,clip]{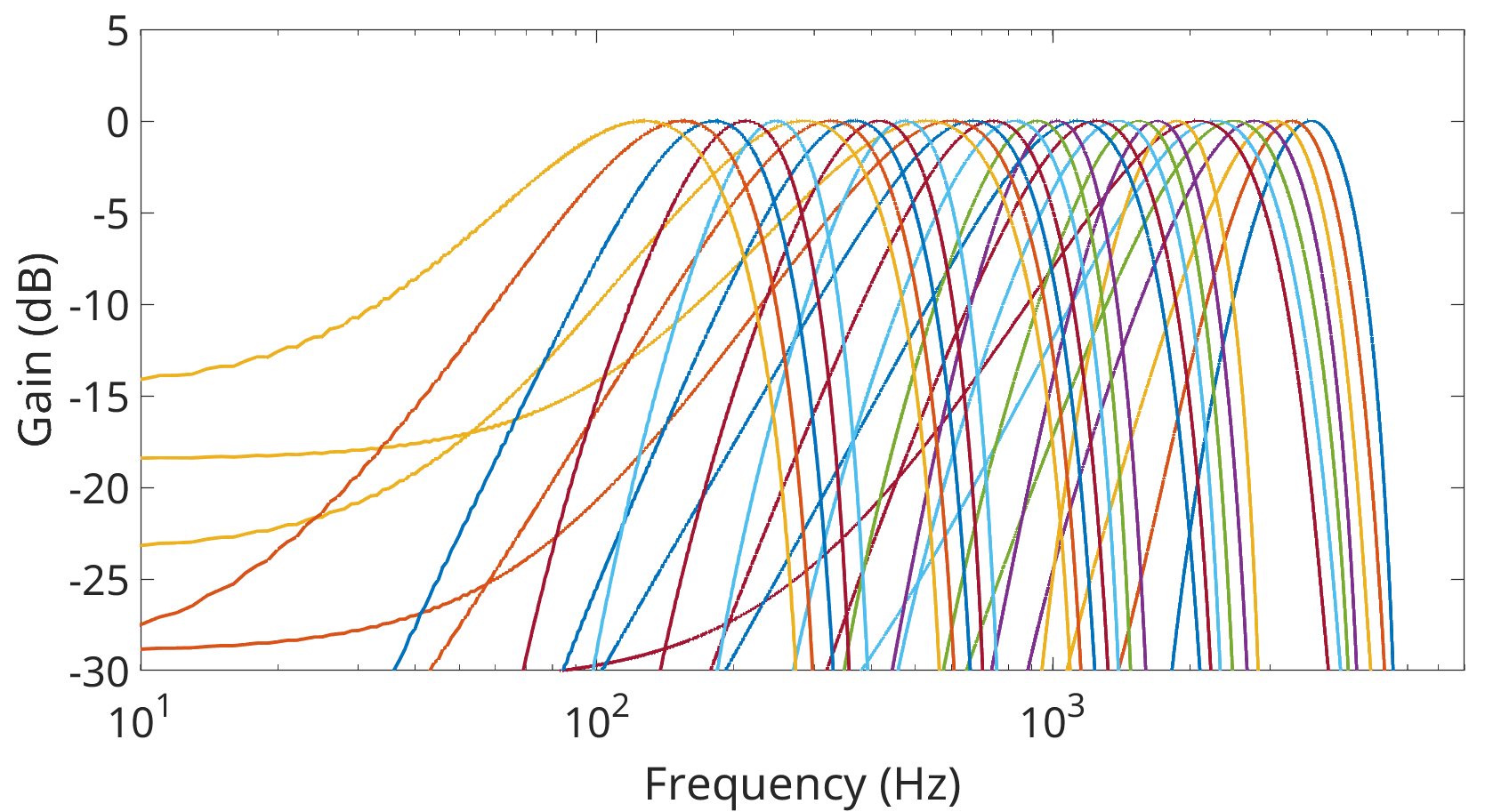}}}
    \subfloat{\label{Fig:Down}\stackinset{l}{.01in}{t}{.05in}{(b)}{\includegraphics[page=1,scale=0.15,trim=0 0 0 0,clip]{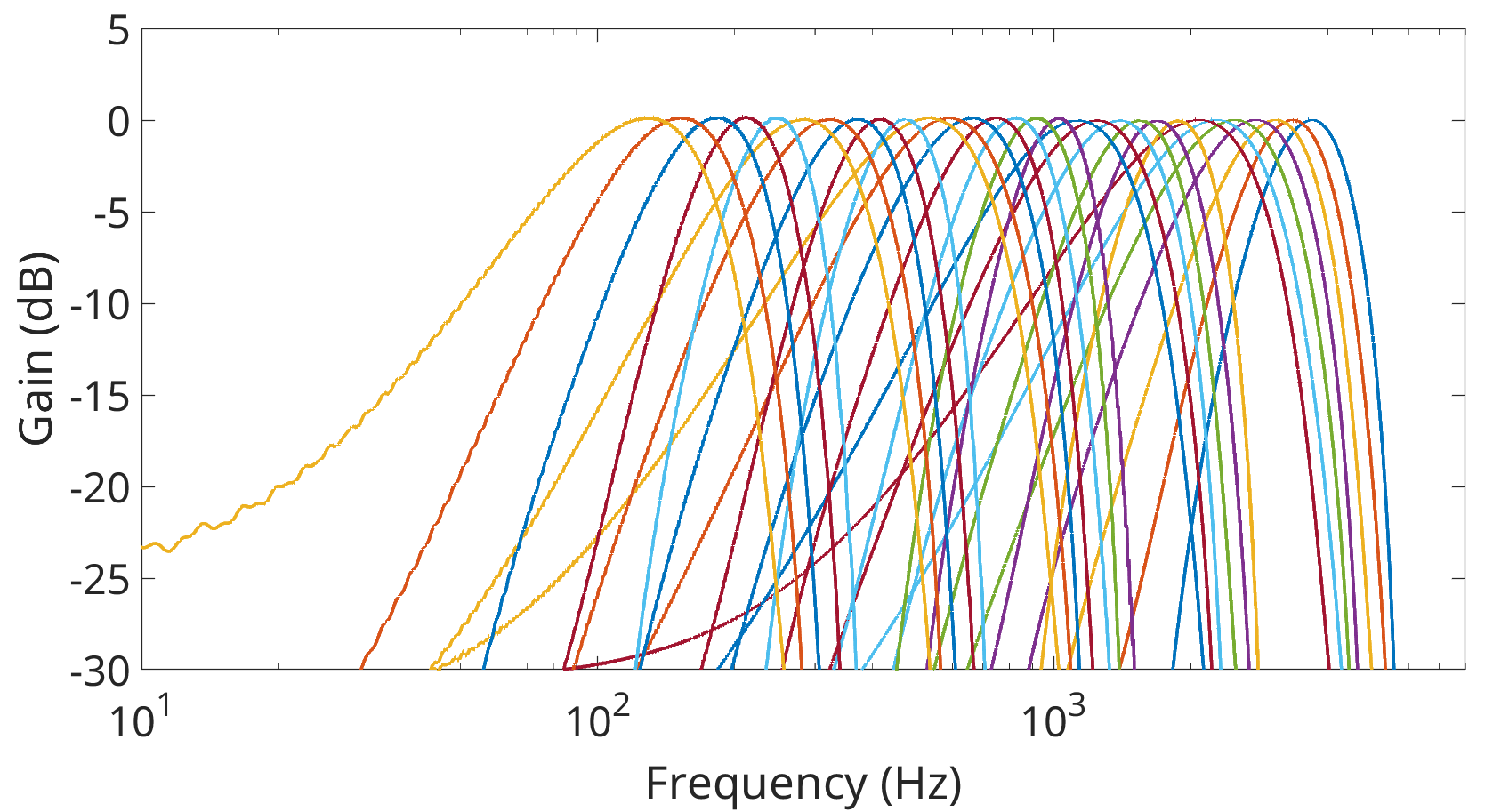}}}
    \caption{FIR Filter Bank Output (Gain Response) for chirp signal (a) Without Downsampling (Filter Order ranges from 15 to 200) (b) With Downsampling (Filter Order fixed at 15)} \label{Fig:chircomp}
\end{figure*}

\subsection{Multiplierless Kernel Machine using MP}\label{kernelmachine}
We develop a classification framework based on multiplierless kernel machine using the MP approximation \cite{nair2022multiplierless}. 
Consider a vector ${\x} \in \mathbb{R}^d$, the decision function for kernel machines~\cite{cristianini2000introduction} is given as,
\begin{align}
    f({\x}) = {\w}^T{\K} + \mathbf{b}. \label{eq:1}
\end{align}
\added{Here, $\K$ is a function of $\x$. }
Following the derivations in ~\cite{nair2022multiplierless}, we can rewrite eq.\eqref{eq:1} in MP domain as,
\begin{align}
    f_{MP}(\x) = z^+ - z^-. \label{eq:2}
\end{align}
where,
\begin{align}
    z^+ = MP([\w^+ + \K^+, \w^- + \K^-, \mathbf{b^+} ],\gamma_1). \label{eq:3} \\
    z^- = MP([\w^+ + \K^-, \w^- + \K^+, \mathbf{b^-} ],\gamma_1). \label{eq:4}
\end{align}
$\gamma_1$ is a hyper-parameter that is learned using gamma annealing. Here $\K^{+} = \K$ and $\K^{-} = -\K$. $\K$ is the kernel which we derive using in-filter computation described in Section \ref{KernelFIR}.
We normalize the values for $z^+$ and $z^-$ for better stability of the system using MP,
\begin {align}
    z = MP([z^+,z^-],\gamma_n). \label{eq:5}
\end{align}
Here, $\gamma_n$ is the hyper-parameter used for normalization. In this case, $\gamma_n = 1$.
The output of the system can be expressed in differential form,
\begin{align}
    p = p^{+} - p^{-}. \label{eq:6}
\end{align}
Here, $p \in \mathbb{R}$, $p^{+} + p^{-} = 1$ and $p^+,p^- \geq 0$.
As $z$ is the normalizing factor for $z^+$ and $z^-$, we can estimate the output using reverse water filling algorithm \cite{gu2012theory}, which is generated by the MP function for each class,
\begin{align}
    p^{+} = [z^{+} - z]_{+}. \nonumber \\
    p^{-} = [z^{-} - z]_{+}. \label{eq:7}
\end{align}

As shown in the Fig.\ref{Fig:Logic_Block_Diag}, the kernel function forms a vector (p $\times$ 1) defined as $\K$.
Using the principle of template based classification described in \cite{kumar2019neuromorphic} and \cite{NairTemplate}, we use the parallel FIR filterbank as the kernel as well as the feature extractor.

\begin{figure}[ht]
\centerline{\includegraphics[page=1,scale=0.5,trim=0 0 1 0,clip]{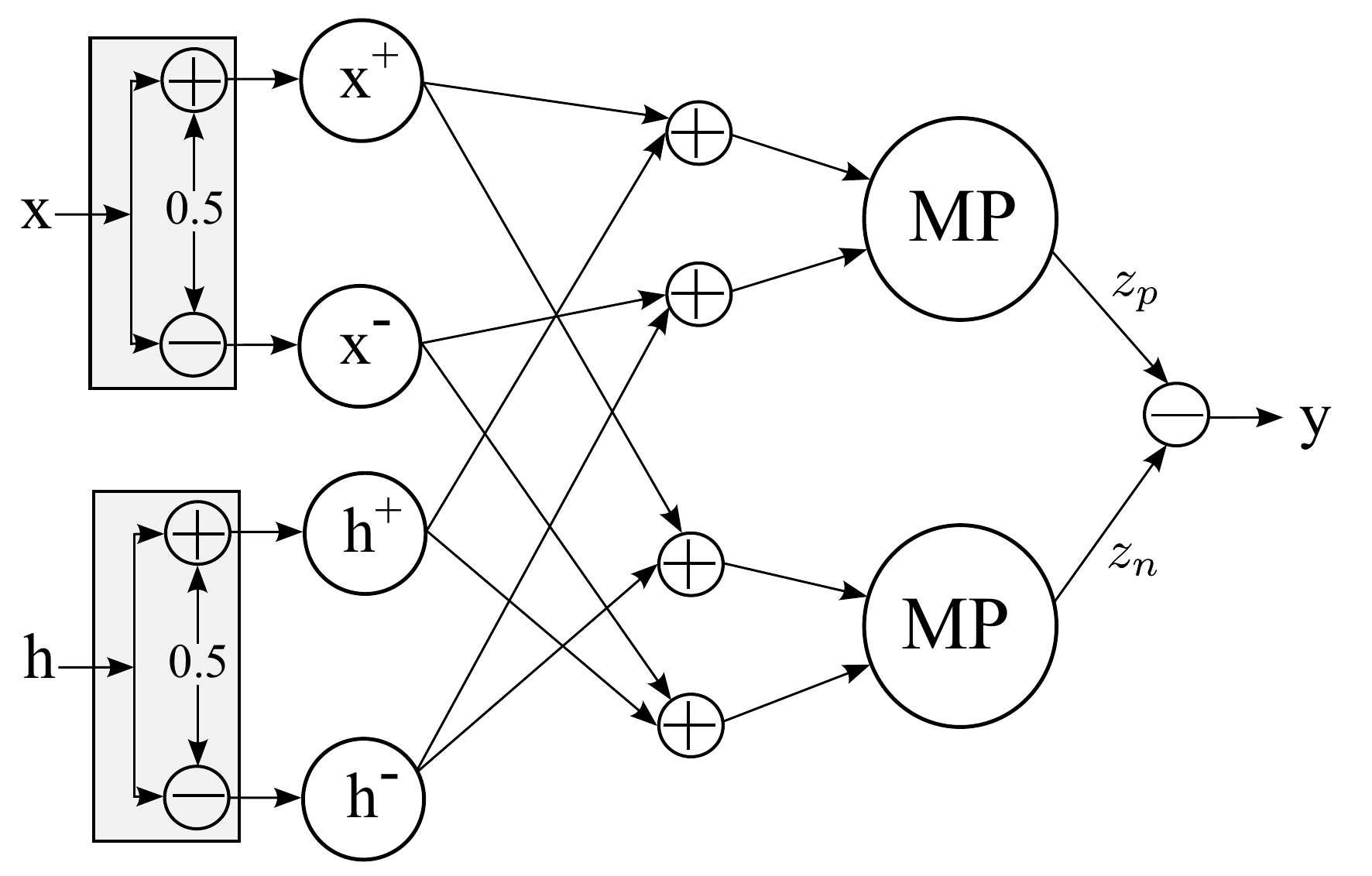}}
\caption{Filtering operation using MP.}
\label{Fig:MPFilter}
\end{figure}

\begin{figure}[ht]
\centerline{\includegraphics[page=1,scale=0.15,trim=1 0 0 0,clip]{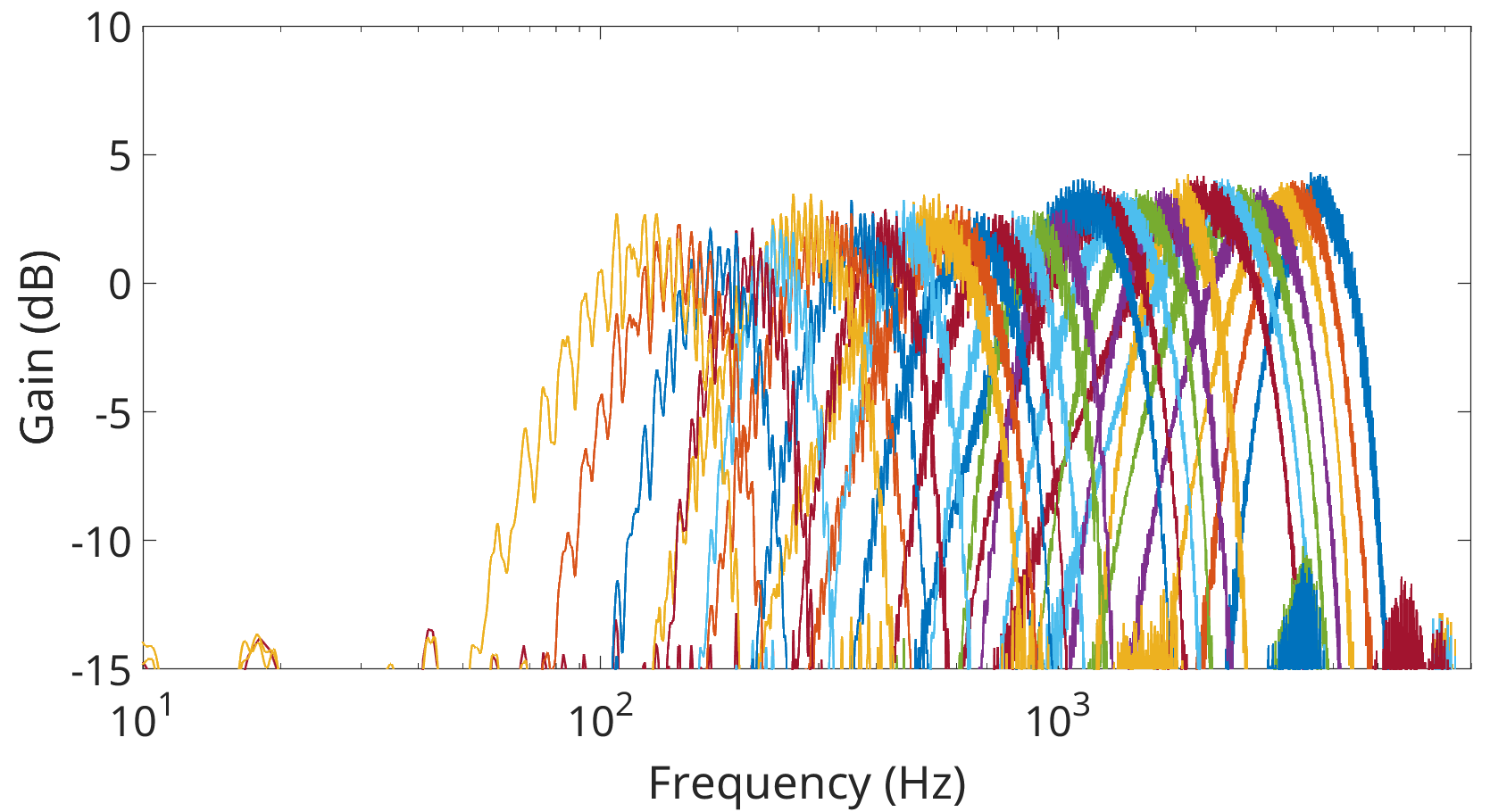}}
\caption{MP Filter bank output (Gain Response) for chirp signal. \added{This response shows distortion due to MP approximation errors.}}
\label{Fig:chirpMP}
\end{figure}

\begin{figure*}[ht]
\centerline{\includegraphics[page=1,scale=0.38,trim=0 0 0 0,clip]{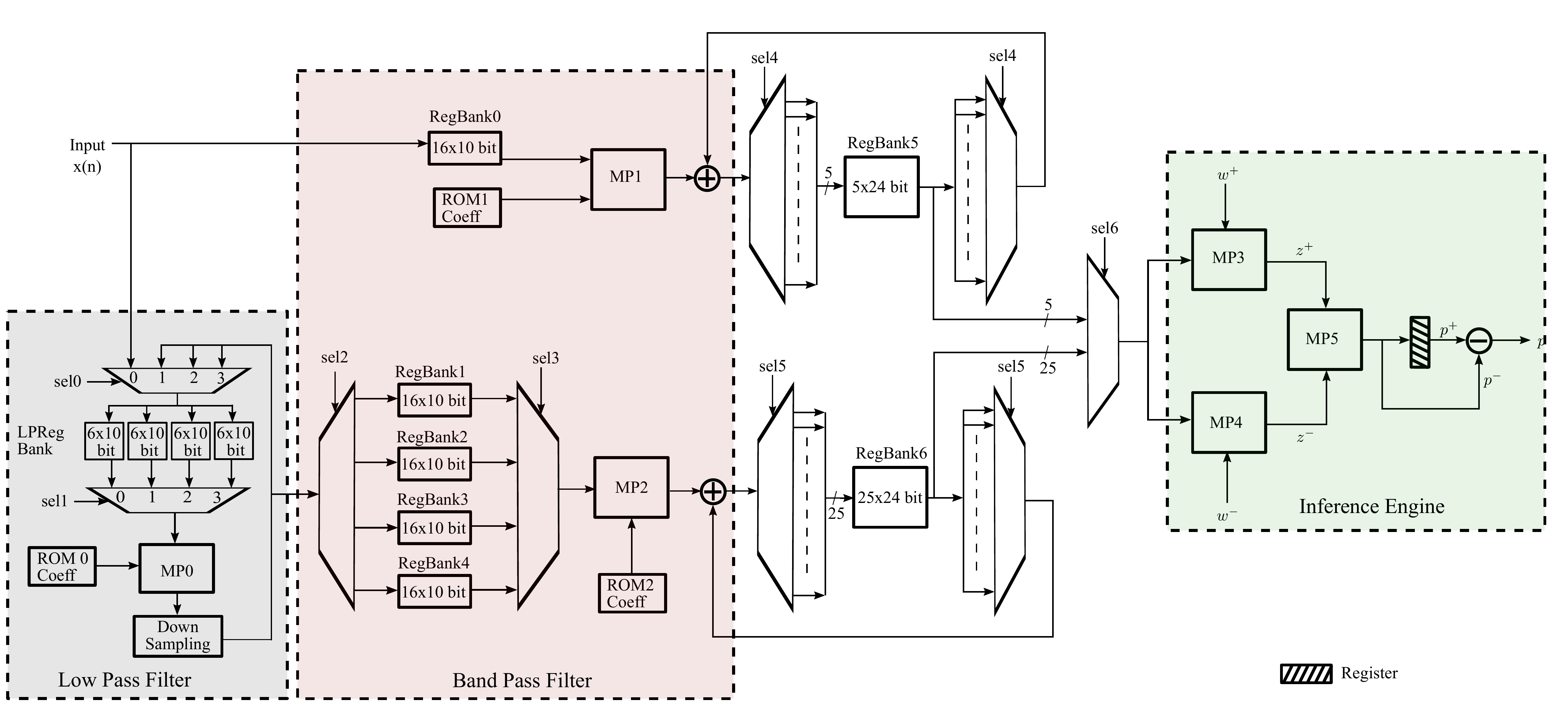}}
\caption{ Architectural implementation of proposed Multiplierless In-Filter Acoustic Classification Framework.}
\label{Fig:HighLevel}
\end{figure*}


\subsection{FIR Filter Bank as Kernel} \label{KernelFIR}
Filter banks are commonly used for feature extraction in acoustic classification \cite{sailor2017unsupervised} \cite{xu2018fpga}. We use FIR filter bank due to it's stability and ease of implementation especially in approximate computing \cite{lu2000design} \cite{soares2015approximate}. Each filter in the filter bank has resonators with center frequencies based on the Greenwood function \cite{greenwood1990cochlear}. Fig.\ref{Fig:Logic_Block_Diag} shows the detailed block architecture of the filter bank. The input $x(n) \in \x$ is an acoustic instance sampled at 16 kHz frequency, i.e, $N = 16000$. $\mathbf{B_p}$ denotes the $p^{th}$ bandpass FIR filter and $p \in P$, i.e., $P$ is the total number of filters in the filter bank.
\begin{align}
    \mathbf{B_p}(n) = \sum\limits_{k=0}^{M-1} h_p(k)x(n-k). \label{eq:8}
\end{align}
Here, $h_p$ is the filter coefficient based on $p^{th}$ filter cut off frequency. $M$ is the order of the filter. The output of the band pass filter is half wave rectified using HWR block, which is then accumulated over $N$ samples and then standardized (STD block) to get the kernel function $\Phi_p \in \mathbb{R}$.
The derivation for this kernel function is described in Appendix \ref{Appendix:kernel}.

The filter bank has been divided into multiple octaves with a bank size of 5 filters per octave. Octaves are defined based on the sampling frequencies in decreasing order. The cut-off frequencies is equally spaced within the octaves. The coefficients ($h_p(n)$) are precomputed and provided as inputs to each filter. We use the technique of downsampling input sampling frequency and segregating the cut-off frequencies into separate octaves, as shown in Fig.\ref{Fig:Logic_Block_Diag}. The cut-off frequencies are arranged in descending order which helps to reduce input sampling frequency. This is a proven efficient way of implementing a filter bank, as shown in \cite{singh2018car}. The downsampling employs a low pass filter (L) used for anti-aliasing at the input for each octave. Downsampling of input ensures usage of lower order FIR filter to obtain the desired output. This is shown in Fig.\ref{Fig:chircomp} using a comparison of the output of an FIR filter bank with and without downsampling for a size of 30 filters. We use a chirp signal as the input with increasing frequency and sampling rate at 16 kHz. We require a range of higher-order FIR filters to get the desired output in filter banks without downsampling. The higher cut-off frequencies required a filter order of 15, and as the cut-off frequencies were reduced, the order of the filters increased, which resulted in the lowest 5 frequencies requiring 200 ordered filters. In contrast, we see that the same output can be obtained with just 15 ordered FIR filters by downsampling the input signal for each range of cut-off frequencies. In this case, the cut-off frequencies were segregated into \added{6} octaves, corresponding to the sampling frequencies in descending order. The additional low pass filter required the same order as the bandpass filter. Thus, this down-sampling technique provides an efficient way of implementing an FIR filter bank for low-powered devices.

\subsection{Filtering operation in MP domain} \label{MPFiltering}
We use two types of filtering operation in our filter bank, i.e., a low pass filter for downsampling and a bandpass filter. These filtering operations result in an inner product computation between the filter coefficients ($h_p(n) \in \mathbf{h}$) and input samples ($x(n)$) as per eq.\eqref{eq:8}.
Following the derivations in \cite{nair2022multiplierless}, we can express this filtering operation in MP domain as below,
\begin{align}
    y = MP\left( \left[ \mathbf{h}^+ + \x^+,\mathbf{h}^- + \x^- \right],\gamma_f \right) \quad \nonumber \\ 
- MP\left( \left[ \mathbf{h}^+ + \x^-, \mathbf{h}^- + \x^+ \right], \gamma_f \right). \label{eq:9}
\end{align}
For this implementation, we have  $\mathbf{h}^+$ = $\mathbf{h}$, $\mathbf{h}^-$ = $-\mathbf{h}$,  $\mathbf{x}^+$ = $\mathbf{x}$ and $\mathbf{x}^-$ = $-\mathbf{x}$. $\gamma_f$ is the MP parameter for the filtering operation. Fig.\ref{Fig:MPFilter} shows the MP implementation of the filtering operation for the low pass and the bandpass filters.
Since the property of MP inherently exhibits low pass filtering, based on the reverse water filling algorithm described in \cite{gu2012theory}, we require a lower-ordered low pass filter implementation in the case of the MP domain. We can see the frequency response of the filter bank in the MP domain in Fig.\ref{Fig:chirpMP}.

We observe some amount of distortion in the gain response of the chirp signal output. This is due to the MP approximation of the inner product for filtering operation. The learning algorithm can mitigate this approximation error, where the weights will be adjusted, considering the approximation error. MP approximation technique minimizes the error rather than mitigating the approximation itself, improving the system's overall accuracy. This technique requires basic primitives like comparators, shift operators, counters, and adders to implement the system, making it hardware-friendly. 



\begin{table}[b]
\centering
\caption{FPGA Implementation Summary.\label{FPGA_SUM}}
\begin{tabular}{|cccc|}
\hline
\multicolumn{1}{|c|}{\multirow{2}{*}{\textbf{Device}}} & \multicolumn{3}{c|}{\multirow{2}{*}{\begin{tabular}[c]{@{}c@{}}Spartan 7\\ xc7s6cpga196-2\end{tabular}}} \\
\multicolumn{1}{|c|}{}                                 & \multicolumn{3}{c|}{}                                                                                    \\ \hline
\multicolumn{3}{|c|}{\textbf{Frequency}}                                                                                                    & 50 MHz              \\ \hline
\multicolumn{3}{|c|}{\added{\textbf{Dynamic Power}}}                                                                                                        & \added{17 mW }              \\ \hline
\multicolumn{3}{|c|}{\textbf{Slices}}                                                                                                       & 903                 \\ \hline
\multicolumn{1}{|c|}{\textbf{FFs}}                     & \multicolumn{1}{c|}{\textbf{LUTs}}       & \multicolumn{1}{c|}{\textbf{DSP}}       & \textbf{BRAM}       \\ \hline
\multicolumn{1}{|c|}{2376}                             & \multicolumn{1}{c|}{1503}                & \multicolumn{1}{c|}{0}                  & 0                   \\ \hline
\end{tabular}
\end{table}

\begin{table*}[]
\centering
\caption{Comparison of architecture and resource utilization of related work.\label{RLUT}}
 \begin{threeparttable}[t]
\begin{tabular}{|l||c|c|c|c|c|c|}
\hline
\textbf{\begin{tabular}[c]{@{}c@{}}Related Work\end{tabular}} &
  \textbf{\begin{tabular}[c]{@{}c@{}}Mahmoodi, et al.\\ \cite{mahmoodi2011fpga}\end{tabular}} &
  \textbf{\begin{tabular}[c]{@{}c@{}}Cutajar, et al.\\ \cite{cutajar2013hardware}\end{tabular}} &
  \textbf{\begin{tabular}[c]{@{}c@{}}Boujelben, et. al.\\ \cite{boujelben2018efficient}\end{tabular}} &
  \textbf{\begin{tabular}[c]{@{}c@{}}Ramos-Lara et al.\\ \cite{ramos2009svm}\end{tabular}} &
  \textbf{\begin{tabular}[c]{@{}c@{}}Nair et al.\\ \cite{NairTemplate}\end{tabular}} &
  \textbf{This work } \\ \hline  \hline
  
\added{\textbf{Year}}          
& \added{2011}              
& \added{2013}               
& \added{2018}                
& \added{2009}                      
& \added{2021}          
& \added{2022}                                                              
\\ \hline
\textbf{FPGA}                         
& \begin{tabular}[c]{@{}c@{}}Virtex4\\xc4vsx35\end{tabular}                     
& \begin{tabular}[c]{@{}c@{}}Virtex-II \\xc2v3000\end{tabular}                 
& \begin{tabular}[c]{@{}c@{}}Artix-7 \\xc7a100T\end{tabular}                      
& \begin{tabular}[c]{@{}c@{}}Spartan 3 \\xcs2000\end{tabular}                    
& \begin{tabular}[c]{@{}c@{}}Spartan 7 \\xc7s6cpga196\end{tabular}    
& \begin{tabular}[c]{@{}c@{}}Spartan 7 \\xc7s6cpga196\end{tabular}                                  \\ \hline
\textbf{Operating Frequency}          
& 151.286 MHz  
& 42.012 MHz   
& 101.74 MHz  
& 50 MHz  
& 25 MHz
& 50 MHz\tnote{2}
\\ \hline
\textbf{Input Sampling Frequency}     
& NA\tnote{1}             
& 16 kHz                    
& 6 kHz                 
& 8 kHz                       
& 16 kHz        
& 16 kHz                                                                 
\\ \hline
\textbf{Flip Flop}                    
& 11589                   
& 1576                     
& 17074                  
& 5351                        
&   2864         
& 2376                                                                
\\ \hline
\textbf{LUTs}                 
& 9141                    
& 11943                    
& 16563          
& 6785                        
&1517                     
& 1503                                                      
\\ \hline
\textbf{RAM (18 Kb)}                          
& 99                       
& NA\tnote{1}               
& 4                                                 
& NA\tnote{1}                          
& 0   
& 0                                                                           
\\ \hline
\textbf{DSP}                          
& 81                        
& 64                      
& 87                                               
& 21                          
& 4             
& 0                                                                
\\ \hline
\added{\textbf{Power (mW/MHz)}}                          
& \added{NA\tnote{1}}                       
& \added{NA\tnote{1}}                      
& \added{1.12}                                               
& \added{NA\tnote{1}}                          
& \added{0.32}             
& \added{0.34}                                                                
\\ \hline
\added{\textbf{Techniques}}                         
& \begin{tabular}[c]{@{}c@{}}\added{SVM}\end{tabular}                     
& \begin{tabular}[c]{@{}c@{}}\added{DWT and }\\\added{SVM}\end{tabular}                 
& \begin{tabular}[c]{@{}c@{}}\added{MFCC and }\\\added{SVM}\end{tabular}                      
& \begin{tabular}[c]{@{}c@{}}\added{FFT and }\\\added{SVM}\end{tabular}                    
& \begin{tabular}[c]{@{}c@{}}\added{CAR-IHC IIR  }\\\added{and SVM}\end{tabular}    
& \begin{tabular}[c]{@{}c@{}}\added{FIR and }\\\added{Kernel Machine} \end{tabular}                 \\ \hline
\added{\textbf{Datasets} }                        
& \begin{tabular}[c]{@{}c@{}} \added{Persian}\\\added{Digits \cite{khosravi2007introducing}}\end{tabular}                     
& \begin{tabular}[c]{@{}c@{}}\added{TIMIT} \\\added{Corpus \cite{garofolo1993darpa}}\end{tabular}   & \begin{tabular}[c]{@{}c@{}}\added{Respiratory}  \\\added{Sound \cite{boujelben2018efficient}}\end{tabular}                      
& \begin{tabular}[c]{@{}c@{}}\added{Speaker} \\ \added{Verification \cite{ramos2009svm}}\end{tabular}                    
& \begin{tabular}[c]{@{}c@{}}\added{ESC-10 and }\\\added{FSDD \cite{piczak2015dataset}}\end{tabular}
& \begin{tabular}[c]{@{}c@{}}\added{ESC-10 and }\\\added{FSDD \cite{piczak2015dataset}}\end{tabular}
\\ \hline
\textbf{\added{Average Accuracy (\%)}}                          
& \added{98}                        
& \added{61}                      
& \added{94}                                               
& \added{95}                          
& \added{88}             
& \added{88}                                                                
\\ \hline
\end{tabular}
\begin{tablenotes}
     \item[1] These works did not report this entity for their designs. 
     \item[2] Maximum operating frequency of the proposed design is 166 MHz.
\end{tablenotes}
\end{threeparttable}%

\end{table*}

\section{FPGA Implementation}\label{FPGA_Impl}

The high-level block diagram of the proposed design is shown in Fig \ref{Fig:HighLevel}. \added{The target frequency of the proposed design is set to 50 MHz, and the input sampling rate is set to 16 KHz. The number of clock cycles available in between two samples are 3125. The architecture is designed such a way that processing of a new sample is completed within this time limit.} Here, 3 MP modules (MP0-2) work simultaneously to meet the time constraints. \added{The MP0 is used to implement 4 low pass (LP) filters and two MP modules (MP1 and MP2) are responsible for Band Pass (BP) filtering operation. The internal architecture and working principle of a MP module is described in \cite{nair2022multiplierless}. The window size of LP filter is 6 and the samples are stored in a register bank of dimension 6\time10-bit. In LP filter section, four register banks (LPRegBank) are used to store inputs for four LP filters. The selection of a particular register bank is done by the select lines $sel0$ and $sel1$. The ROM (ROM0) is used to store coefficients for LP filters. The precision of the data path  is set to 10 bits for the proposed design.} 
Initially, the input samples ($x(n)$) are stored in a register bank and fed to the MP0 for implementing LP filter $L_1$, and the output of $L_1$ is down-sampled by 2 and passed to the corresponding parallel BP filter bank (for generating octave 2) (as discussed in Section. \ref{KernelFIR}). \added{Here, MP0 implements 4 LP filters in time multiplexed fashion and generates desired outputs for Octave 1, 2, 3 and 4. The outputs are  stored in corresponding register banks (RegBank1-4) using select signal $sel1$. The contents of the register banks (RegBank1-4) are used for parallel BP filtering operation and generates kernel function $\Phi_5$ to $\Phi_{30}$.} 

\added{One single MP module (MP1) is used repeatedly to generate outputs for octave 1. The window size of the BP filter is 16. The inputs and coefficients are stored in register bank (RegBank0) and ROM (ROM1) respectively. The register bank RegBank5, is used to store accumulated values of filter outputs of all 16000 samples i.e., when a output is generated from a filter it added with the previous value and stored in corresponding register in RegBank5. The MP2 is used for BP filter outputs of octaves 2,3,4 and 5. Here also a single MP module used repeatedly to generate desired outputs. The down sampling of the LP filter outputs provides more time span between two outputs which are the inputs to the BP filters generating octaves 2,3,4 and 5. Hence a single MP (MP2) is sufficient to produce $\Phi_5$ to $\Phi_{30}$ in time multiplexed fashion. Like RegBank5, the RegBank6 is also used to store accumulated values of filter outputs of all 16000 samples. The select lines $sel4$ and $sel5$ are used to access a particular register in the register banks RegBank5 and RegBank6 respectively. The contents of the RegBank5 and RegBank6 are the kernel function $\Phi_1$ to $\Phi_{30}$ of the proposed design. The ROM2 is used to store coefficients of BP filters for octaves 2, 3, 4 and 5.}

\begin{table*}[ht]
\centering

\caption{ESC-10 dataset classification accuracy results in percent. Number of filters for our work is fixed at 30. We used 8-bit fixed point for our design  \label{Table:ESC10}}
\begin{tabular}{|c||c|c|c|c|c|c|c|c|c|}
\hline
\multirow{3}{*}{\textbf{Classes}}                          & \multicolumn{3}{c|}{\textbf{Normal SVM}}                                                                  & \multicolumn{2}{c|}{\textbf{CARIHC SVM}}                                  & \multicolumn{4}{c|}{\textbf{MP In-Filter Compute}}                                                                                           \\ \cline{2-10} 
                                                  & \multicolumn{3}{c|}{\textbf{Floating Point}}                                                              & \multicolumn{2}{c|}{\textbf{Floating Point}}                              & \multicolumn{2}{c|}{\textbf{Floating Point}}                              & \multicolumn{2}{c|}{\textbf{Fixed Point (8-bit)}}                         \\ \cline{2-10} 
                                                  & \textbf{SVs} & \textbf{Train} & \textbf{Test} & \textbf{Train} & \textbf{Test} & \textbf{Train} & \textbf{Test} & \textbf{Train} & \textbf{Test} \\ \hline \hline
\textbf{Dog (129/33)}            & 42                            & 90                              & 94                             & 89                              & 90                             & 91                           & 94                           & 91                            & 94                           \\ \hline
\textbf{Rain (119/40)}           & 44                            & 86                              & 90                             & 89                              & 87                             & 90                           & 90                             & 88                           & 88                           \\ \hline
\textbf{Sea\_Waves (200/50)}     & 80                            & 87                              & 90                             & 84                              & 78                             & 89                            & 88                             & 88                           & 88                           \\ \hline
\textbf{Crying Baby (144/49)}    & 37                            & 93                              & 84                             & 91                              & 87                             & 92                           & 87                           & 89                           & 88                           \\ \hline
\textbf{Clock Tick (114/50)}     & 54                            & 81                              & 76                             & 92                              & 85                             & 85                           & 86                             & 85                           & 84                             \\ \hline
\textbf{Person Sneeze (101/44)}  & 49                            & 85                              & 75                             & 87                              & 80                             & 86                           & 80                             & 85                           & 80                             \\ \hline
\textbf{Helicopter (197/50)}     & 45                            & 92                              & 88                             & 95                              & 90                             & 88                           & 85                           & 85                           & 86                           \\ \hline
\textbf{Chainsaw (99/34)}        & 41                            & 90                              & 85                             & 93                              & 82                             & 92                           & 81                           & 92                           & 80                           \\ \hline
\textbf{Rooster (124/54)}        & 40                            & 93                              & 93                             & 93                              & 96                             & 90                           & 94                           & 91                           & 94                           \\ \hline
\textbf{Fire Crackling (152/66)} & 46                            & 93                              & 83                             & 89                              & 87                             & 89                           & 92                           & 90                           & 88                           \\ \hline
\end{tabular}
\end{table*}

\begin{table*}[ht]
\centering

\caption{FSDD classification accuracy results in percent. Number of filters for our work is fixed at 30. We used 8-bit fixed point for our design \label{Table:FSDD}}

\begin{tabular}{|c||c|c|c|c|c|c|c|c|c|}
\hline
\multirow{3}{*}{\textbf{Classes}} & \multicolumn{3}{c|}{\textbf{Normal SVM}}      & \multicolumn{2}{c|}{\textbf{CARIHC SVM}}     & \multicolumn{4}{c|}{\textbf{MP Kernel}}                                                  \\ \cline{2-10} 
                                  & \multicolumn{3}{c|}{\textbf{Floating Point}}  & \multicolumn{2}{c|}{\textbf{Floating Point}} & \multicolumn{2}{c|}{\textbf{Floating Point}} & \multicolumn{2}{c|}{\textbf{Fixed Point}} \\ \cline{2-10} 
                                  & \textbf{SVs} & \textbf{Train} & \textbf{Test} & \textbf{Train}        & \textbf{Test}        & \textbf{Train}        & \textbf{Test}        & \textbf{Train}       & \textbf{Test}      \\ \hline \hline
\textbf{Theo (761/254)}           & 107          & 96             & 96            & 93                    & 91                   & 92                    & 93                   & 92                     & 92                    \\ \hline
\textbf{Nicolas(889/297)}         & 15           & 100            & 100           & 98                    & 97                   & 99                    & 99                   &  98                    & 98                   \\ \hline
\end{tabular}
\end{table*}

\added{The inference engine starts working after the completion of kernel computation.The three MP blocks MP3, MP4 and MP5 are used in the inference engine to generate the output $p$.The architecture and working principle of the inference engine are discussed in Section \ref{kernelmachine}. The $w^+$ and $w^-$  are the weight matrix and stored in a ROM. The kernel function $\Phi$ and weight matrix $w^+$ and $w^-$ are the inputs to the inference engine. The kernel function is selected sequentially by select signal $sel6$. The upper 10 bits of the kernel function ($\Phi_1$ to $\phi_{30}$) are used for inference engine. The high-level block diagram of an MP module and the implementation details of the inference engine has been discussed in \cite{nair2022multiplierless}. 

We have used Spartan 7 series FPGA to implement our design, as this FPGA caters to edge applications. Table \ref{FPGA_SUM} shows the resource utilization and power consumption of our design. We were able to implement our design with usage of less than 1K of FPGA slices and the dynamic power consumption is limited to 17 mW for a 50 MHz operating frequency. Table \ref{RLUT} compares similar designs using varied edge datasets for resource utilization and power consumption. We see a better resource utilization of our design in comparison to these systems with lower power consumption in mW/MHz. The proposed study consumes almost the same amount of LUTs and 488 fewer FFs than the similar design presented in \cite{NairTemplate}.} Due to multiplierless design, the proposed architecture does not consume any DSPs, whereas the design reported in \cite{NairTemplate} consumes 4 DSPs. We computed the number of LUTs required to replace 4 DSPs for fair comparisons. We have implemented 4$\times$4, and 8$\times$8 signed multipliers (Baugh Wooley) in FPGA and found that they have consumed 19 and 72 (~4$\times$ more) LUTs, respectively. The design reported in \cite{NairTemplate} uses 4 signed multipliers and the dimensions are 20$\times$12, 20$\times$12, 12$\times$12, 16$\times$8 respectively. The approximation calculation shows that all 4 multipliers consume at least 890 LUTs. Hence the proposed multiplierless design can save at least 25\% hardware resources (LUTs + FFs) compared to the design presented in \cite{NairTemplate}.
\added{The power consumption (mW/MHz) is almost same for the proposed design and the design presented in \cite{NairTemplate}, as the design is small and only 4 multipliers are used. However, for the bigger network such as DNN our multiplierless approach would give significant benefit. The proposed design can achieve maximum operation frequency of 166 MHz which can be used to support more input sampling rate.}


\begin{figure}[ht]
\centerline{\includegraphics[page=1,scale=0.15,trim=0 0 1 0,clip]{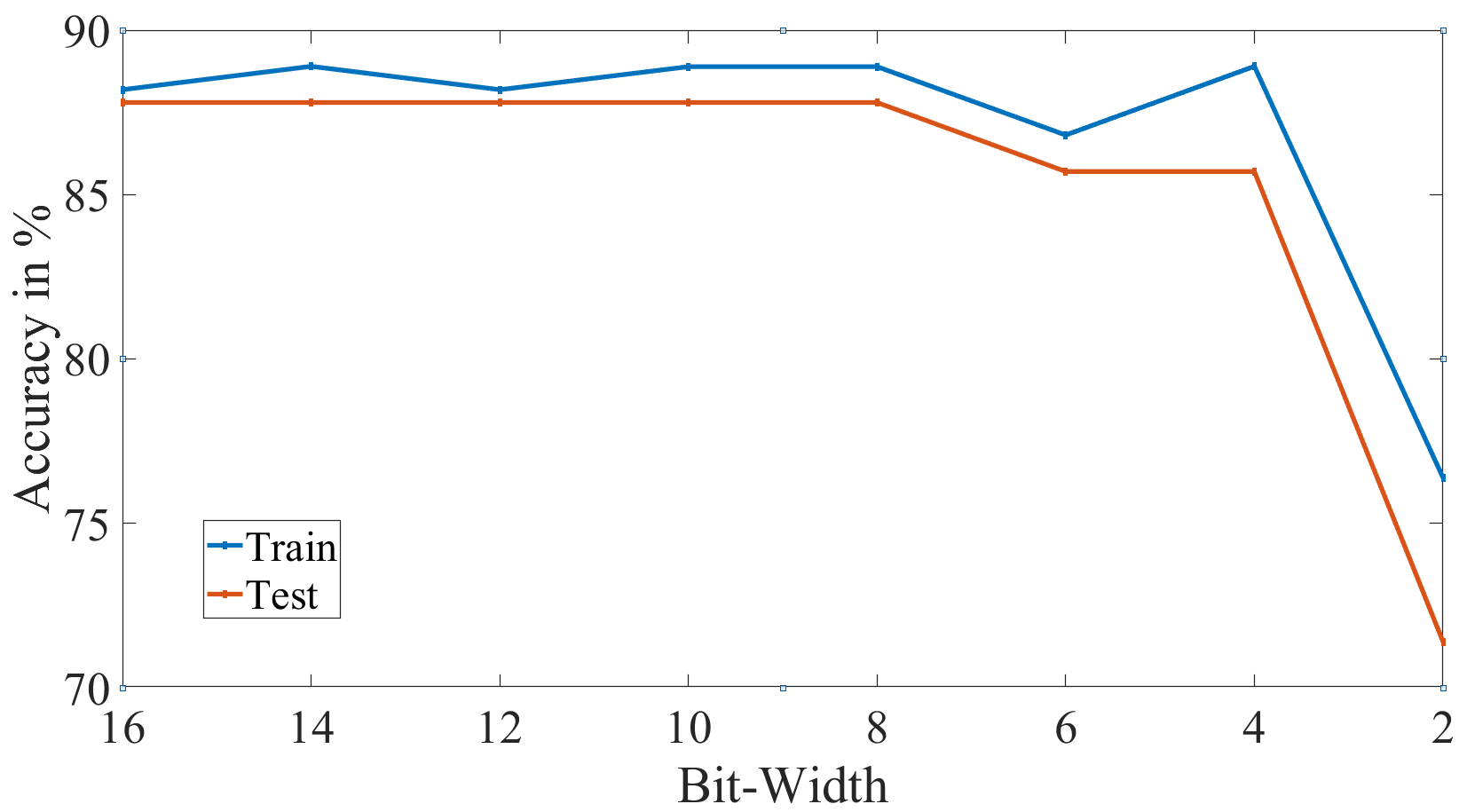}}
\caption{Impact of bit-width on dataset accuracy for Crying Baby class from ESC-10.}
\label{Fig:Dataset_bit}
\end{figure}

\section{Results and Discussion} \label{Results_Discussion}
The framework's classification ability is showcased using the environmental sounds dataset. Identification of different environmental sounds shows the versatile nature of the framework that can be put to use in various acoustic applications. As wildlife conservation would result in rare data event detection, Environmental Sounds Classification (ESC-10) dataset \cite{piczak2015dataset} would be an ideal dataset to showcase the framework capabilities in case of ecological application. Also, we compared our system with \cite{NairTemplate} using ESC-10 and Free Speech Digit Dataset (FSDD), where we use speaker identification as the application.

ESC-10 dataset consists of sound clips constructed from recordings publicly available through the Freesound project. It consists of 400 environmental recordings with 10 classes, i.e., 40 clips per class and 5 seconds per clip. Each class contains 40 wav format audio files. These clips had a lot of silence, so we trimmed the silence part and further trimmed the remaining clips into a 1-second version of the same class, thus increasing the dataset's number of samples. Table \ref{Table:ESC10} shows the results for this dataset, having classes like a dog bark, rain, sea waves, crying baby, clock ticking, person sneezing, helicopter, chainsaw, crawing rooster, and fire crackling. The classification uses one versus all methodology to identify the classes, where the data is balanced and randomly arranged for train and test sets (shown in brackets). We use the in-built MATLAB library with default command lines for the traditional SVM. \added{Here, the CARIHC SVM employs a completely different approach compared to standard SVM for arriving at the accuracy, which is detailed in \cite{NairTemplate}, \cite{kumar2019neuromorphic}. Since the dataset size is small, the accuracy values differ by a bigger margin for some classes between the traditional SVM and the other two SVM implementation, as small variations in positive or negative classification will lead to a greater impact on accuracy number.} Similarly, we compare the identification of two speakers from the FSDD dataset. These results show that our framework takes advantage of template SVM methodology with a fixed number of templates and the MP approximation technique, delivering comparable results. We have also compared our system with similar systems, which is area efficient, as shown in Table \ref{RLUT}. 

We use an 8-bit fixed point for implementing the hardware. We performed an empirical analysis of the dataset (using the crying baby class) with different bit widths. As shown in Fig.\ref{Fig:Dataset_bit}, the training and testing accuracy remains stable till 8-bit and decreases sharply for bit width lower than 8-bit. We use Keras \cite{chollet2015keras} implementation for training our system, as this software framework is robust and highly optimized. The FIR filter banks are quantized at 8-bit, and a custom layer for the MP function is written for the Keras framework. The optimization of the model is done using Tensorflow \cite{abadi2016tensorflow} libraries for quantization. 


\section{Conclusion}
This paper presents a novel multiplierless framework for acoustic classification using an FIR filterbank as the feature extraction and kernel function stage simultaneously. This framework is entirely multiplierless since the FIR filter bank is implemented using MP approximation along with the inference logic. Furthermore, the framework is tunable to any time series data by tuning the filter parameters in the FIR filter bank. The framework's scalability is evident as the number of filters is user-defined and can be controlled to adhere to IoT system constraints. Being completely multiplierless makes the system highly efficient for deployment in battery-powered edge devices. Various time series data generated from different biosensors can be used in their raw format and for classification using this framework without additional preprocessing or hardware. A network of edge devices running our proposed classification framework can be used for continuous monitoring of wildlife species and detecting anomalies in case of poaching or timber smuggling. This framework can be extended to other biomedical applications using edge devices capable of healthcare monitoring with raw ECG, EMG, and EEG signals. Wearable IMU sensors with this framework can be used to detect anomalies in posture. To make our framework more energy efficient, we can fabricate this system into an Application Specific Integrated Chip.


\section*{Appendix}

\renewcommand{\thesubsection}{\Alph{subsection}}

\subsection{FIR Filter Kernel Derivation} \label{Appendix:kernel}

The output of each band pass filter, as shown in Fig.\ref{Fig:Logic_Block_Diag}, forms the input to the HWR blocks in parallel.
\begin{equation}
    HWR(q) = max(0,q). \label{eq:a1}
\end{equation}

The half wave rectified output is summed over $N$ samples of a single input, and this forms the input for the standardization ($STD$) blocks in parallel.  
Let $d_p(n) = HWR_p(B_p(n))$,
\begin{equation}
    s_p = \sum\limits_{n=1}^{N}d_p(n). \label{eq:a2}
\end{equation}
Here, $s_p \in \mathbb{R}$.
\begin{equation}
   STD(S_{p,i})=\frac{S_{p,i}-\mu_p}{\sigma_p}.  \label{eq:a3} 
\end{equation}
where $\{ s_p \in S_{p,i} | 1 \leq i \leq M \}$ with $M$ as the number of training inputs, $\mu_p = mean(S_{p,1}, S_{p,2}, .., S_{p,M})$ and  \\ $\sigma_p$ =$\sqrt{\frac{1}{M-1}\sum\limits_{i=1}^M(S_{p,i}-\mu_p)^2}$

\begin{equation}
   \Phi_p = STD(S_{p,i}).\label{eq:a4}
\end{equation}
Here, $\Phi_p  \in  \mathbb{R} $.

The summation over $N$ samples of the output of HWR is taken as per eq.\eqref{eq:a2} for each filter. Then  standardization technique, commonly used in neural network optimizations ~\cite{shanker1996effect}, is applied  across $M$ training input samples as per eq.\eqref{eq:a3}. Note that $\mu_p$ and $\sigma_p$ are calculated only during training, and these vectors are passed as learned parameters to the inference engine. 
Therefore, an input signal vector $X_n$ sampled at a sampling frequency $f_{s}$ generates $N$ samples with each sample denoted as $x(n)$. It is then processed by a parallel FIR filter bank to estimate the kernel function $\Phi_p$ with ${p}$ as the filter stage out of $P$ filters as per \eqref{eq:a4}. 
The output is a $P\times 1$ kernel vector ($\K$), as shown in Fig.\ref{Fig:Logic_Block_Diag}.

\section*{Acknowledgment}
This research was supported in part by (i) INSPIRE faculty fellowship 
(DST/INSPIRE/04/2016/000216), SPARC grant (SPARC/2018-2019/P606/SL) from Ministry of Human Resource Development and IMPRINT Grant IMP/2018/000550 from the Department of Science 
and Technology, India. 
The authors would like to acknowledge the joint Memorandum of Understanding (MoU) between Indian Institute of Science, Bangalore and Washington University in St. Louis for supporting this research activity.

\ifCLASSOPTIONcaptionsoff
  \newpage
\fi



%
\medskip


\bibliographystyle{IEEEtran}
\bibliography{references}

\begin{thebibliography}{10}
\providecommand{\url}[1]{#1}
\csname url@samestyle\endcsname
\providecommand{\newblock}{\relax}
\providecommand{\bibinfo}[2]{#2}
\providecommand{\BIBentrySTDinterwordspacing}{\spaceskip=0pt\relax}
\providecommand{\BIBentryALTinterwordstretchfactor}{4}
\providecommand{\BIBentryALTinterwordspacing}{\spaceskip=\fontdimen2\font plus
\BIBentryALTinterwordstretchfactor\fontdimen3\font minus
  \fontdimen4\font\relax}
\providecommand{\BIBforeignlanguage}[2]{{%
\expandafter\ifx\csname l@#1\endcsname\relax
\typeout{** WARNING: IEEEtran.bst: No hyphenation pattern has been}%
\typeout{** loaded for the language `#1'. Using the pattern for}%
\typeout{** the default language instead.}%
\else
\language=\csname l@#1\endcsname
\fi
#2}}
\providecommand{\BIBdecl}{\relax}
\BIBdecl

\bibitem{tobore2019deep}
I.~Tobore, J.~Li, L.~Yuhang, Y.~Al-Handarish, A.~Kandwal, Z.~Nie, L.~Wang
  \emph{et~al.}, ``Deep learning intervention for health care challenges: some
  biomedical domain considerations,'' \emph{JMIR mHealth and uHealth}, vol.~7,
  no.~8, p. e11966, 2019.

\bibitem{zhao2020role}
B.~Zhao, J.~Mao, J.~Zhao, H.~Yang, and Y.~Lian, ``The role and challenges of
  body channel communication in wearable flexible electronics,'' \emph{IEEE
  Transactions on Biomedical Circuits and Systems}, vol.~14, no.~2, pp.
  283--296, 2020.

\bibitem{witmer2005wildlife}
G.~W. Witmer, ``Wildlife population monitoring: some practical
  considerations,'' \emph{Wildlife Research}, vol.~32, no.~3, pp. 259--263,
  2005.

\bibitem{zgank2021iot}
A.~Zgank, ``Iot-based bee swarm activity acoustic classification using deep
  neural networks,'' \emph{Sensors}, vol.~21, no.~3, p. 676, 2021.

\bibitem{quintana2017low}
M.~A. Quintana-Su{\'a}rez, D.~S{\'a}nchez-Rodr{\'\i}guez,
  I.~Alonso-Gonz{\'a}lez, and J.~B. Alonso-Hern{\'a}ndez, ``A low cost wireless
  acoustic sensor for ambient assisted living systems,'' \emph{Applied
  Sciences}, vol.~7, no.~9, p. 877, 2017.

\bibitem{NairTemplate}
A.~R. Nair, S.~Chakrabartty, and C.~S. Thakur, ``In-filter computing for
  designing ultra-light acoustic pattern recognizers,'' \emph{IEEE Internet of
  Things Journal}, 2021.

\bibitem{popovic2017architecting}
T.~Popovi{\'c}, N.~Latinovi{\'c}, A.~Pe{\v{s}}i{\'c},
  {\v{Z}}.~Ze{\v{c}}evi{\'c}, B.~Krstaji{\'c}, and S.~Djukanovi{\'c},
  ``Architecting an iot-enabled platform for precision agriculture and
  ecological monitoring: A case study,'' \emph{Computers and electronics in
  agriculture}, vol. 140, pp. 255--265, 2017.

\bibitem{luo2021localization}
J.~Luo, Y.~Yang, Z.~Wang, and Y.~Chen, ``Localization algorithm for underwater
  sensor network: A review,'' \emph{IEEE Internet of Things Journal}, 2021.

\bibitem{de2013wireless}
A.~De~La~Piedra, F.~Benitez-Capistros, F.~Dominguez, and A.~Touhafi, ``Wireless
  sensor networks for environmental research: A survey on limitations and
  challenges,'' in \emph{Eurocon 2013}.\hskip 1em plus 0.5em minus 0.4em\relax
  IEEE, 2013, pp. 267--274.

\bibitem{tuia2022perspectives}
D.~Tuia, B.~Kellenberger, S.~Beery, B.~R. Costelloe, S.~Zuffi, B.~Risse,
  A.~Mathis, M.~W. Mathis, F.~van Langevelde, T.~Burghardt \emph{et~al.},
  ``Perspectives in machine learning for wildlife conservation,'' \emph{Nature
  communications}, vol.~13, no.~1, pp. 1--15, 2022.

\bibitem{shan2006machine}
Y.~Shan, D.~Paull, and R.~McKay, ``Machine learning of poorly predictable
  ecological data,'' \emph{Ecological Modelling}, vol. 195, no. 1-2, pp.
  129--138, 2006.

\bibitem{heydari2018effect}
S.~S. Heydari and G.~Mountrakis, ``Effect of classifier selection, reference
  sample size, reference class distribution and scene heterogeneity in
  per-pixel classification accuracy using 26 landsat sites,'' \emph{Remote
  Sensing of Environment}, vol. 204, pp. 648--658, 2018.

\bibitem{han2015deep}
S.~Han, H.~Mao, and W.~J. Dally, ``Deep compression: Compressing deep neural
  networks with pruning, trained quantization and huffman coding,'' \emph{arXiv
  preprint arXiv:1510.00149}, 2015.

\bibitem{tang2008svms}
Y.~Tang, Y.-Q. Zhang, N.~V. Chawla, and S.~Krasser, ``Svms modeling for highly
  imbalanced classification,'' \emph{IEEE Transactions on Systems, Man, and
  Cybernetics, Part B (Cybernetics)}, vol.~39, no.~1, pp. 281--288, 2008.

\bibitem{nair2019hybrid}
P.~Nair and I.~Kashyap, ``Hybrid pre-processing technique for handling
  imbalanced data and detecting outliers for knn classifier,'' in \emph{2019
  International Conference on Machine Learning, Big Data, Cloud and Parallel
  Computing (COMITCon)}.\hskip 1em plus 0.5em minus 0.4em\relax IEEE, 2019, pp.
  460--464.

\bibitem{maalouf2011robust}
M.~Maalouf and T.~B. Trafalis, ``Robust weighted kernel logistic regression in
  imbalanced and rare events data,'' \emph{Computational Statistics \& Data
  Analysis}, vol.~55, no.~1, pp. 168--183, 2011.

\bibitem{horowitz20141}
M.~Horowitz, ``1.1 computing's energy problem (and what we can do about it),''
  in \emph{2014 IEEE International Solid-State Circuits Conference Digest of
  Technical Papers (ISSCC)}.\hskip 1em plus 0.5em minus 0.4em\relax IEEE, 2014,
  pp. 10--14.

\bibitem{guo2018survey}
Y.~Guo, ``A survey on methods and theories of quantized neural networks,''
  \emph{arXiv preprint arXiv:1808.04752}, 2018.

\bibitem{tagliavini2018transprecision}
G.~Tagliavini, S.~Mach, D.~Rossi, A.~Marongiu, and L.~Benini, ``A
  transprecision floating-point platform for ultra-low power computing,'' in
  \emph{2018 Design, Automation \& Test in Europe Conference \& Exhibition
  (DATE)}.\hskip 1em plus 0.5em minus 0.4em\relax IEEE, 2018, pp. 1051--1056.

\bibitem{hubara2017quantized}
I.~Hubara, M.~Courbariaux, D.~Soudry, R.~El-Yaniv, and Y.~Bengio, ``Quantized
  neural networks: Training neural networks with low precision weights and
  activations,'' \emph{The Journal of Machine Learning Research}, vol.~18,
  no.~1, pp. 6869--6898, 2017.

\bibitem{fan2020training}
A.~Fan, P.~Stock, B.~Graham, E.~Grave, R.~Gribonval, H.~Jegou, and A.~Joulin,
  ``Training with quantization noise for extreme model compression,''
  \emph{arXiv preprint arXiv:2004.07320}, 2020.

\bibitem{rastegari2016xnor}
M.~Rastegari, V.~Ordonez, J.~Redmon, and A.~Farhadi, ``Xnor-net: Imagenet
  classification using binary convolutional neural networks,'' in
  \emph{European conference on computer vision}.\hskip 1em plus 0.5em minus
  0.4em\relax Springer, 2016, pp. 525--542.

\bibitem{zhou2016dorefa}
S.~Zhou, Y.~Wu, Z.~Ni, X.~Zhou, H.~Wen, and Y.~Zou, ``Dorefa-net: Training low
  bitwidth convolutional neural networks with low bitwidth gradients,''
  \emph{arXiv preprint arXiv:1606.06160}, 2016.

\bibitem{hemanth2022}
H.~Sabbella, A.~R. Nair, V.~Gumme, S.~Yadav, S.~Chakrabartty, and C.~S. Thakur,
  ``An always-on tinyml acoustic classifier for ecological applications,'' in
  \emph{ISCAS}, 2022.

\bibitem{kumar2019neuromorphic}
P.~Kumar, A.~R. Nair, O.~Chatterjee, T.~Paul, A.~Ghosh, S.~Chakrabartty, and
  C.~S. Thakur, ``Neuromorphic in-memory computing framework using
  memtransistor cross-bar based support vector machines,'' in \emph{2019 IEEE
  62nd International Midwest Symposium on Circuits and Systems (MWSCAS)}.\hskip
  1em plus 0.5em minus 0.4em\relax IEEE, 2019, pp. 311--314.

\bibitem{chakrabartty2004margin}
S.~Chakrabartty and G.~Cauwenberghs, ``Margin propagation and forward decoding
  in analog vlsi,'' \emph{A (A)}, vol. 100, p.~5, 2004.

\bibitem{nair2022multiplierless}
A.~R. Nair, P.~K. Nath, S.~Chakrabartty, and C.~S. Thakur, ``Multiplierless
  mp-kernel machine for energy-efficient edge devices,'' \emph{IEEE
  Transactions on Very Large Scale Integration (VLSI) Systems}, 2022.

\bibitem{singh2018car}
R.~K. Singh, Y.~Xu, R.~Wang, T.~J. Hamilton, A.~van Schaik, and S.~L. Denham,
  ``Car-lite: A multi-rate cochlea model on fpga,'' in \emph{2018 IEEE
  International Symposium on Circuits and Systems (ISCAS)}.\hskip 1em plus
  0.5em minus 0.4em\relax IEEE, 2018, pp. 1--5.

\bibitem{piczak2015dataset}
K.~J. Piczak, ``{ESC}: {Dataset} for {Environmental Sound Classification},''
  \emph{Proceedings of the 23rd {Annual ACM Conference} on {Multimedia}}, pp.
  1015--1018, 2015.

\bibitem{nguyen2017animal}
H.~Nguyen, S.~J. Maclagan, T.~D. Nguyen, T.~Nguyen, P.~Flemons, K.~Andrews,
  E.~G. Ritchie, and D.~Phung, ``Animal recognition and identification with
  deep convolutional neural networks for automated wildlife monitoring,'' in
  \emph{2017 IEEE international conference on data science and advanced
  Analytics (DSAA)}.\hskip 1em plus 0.5em minus 0.4em\relax IEEE, 2017, pp.
  40--49.

\bibitem{weninger2011audio}
F.~Weninger and B.~Schuller, ``Audio recognition in the wild: Static and
  dynamic classification on a real-world database of animal vocalizations,'' in
  \emph{2011 IEEE International Conference on Acoustics, Speech and Signal
  Processing (ICASSP)}.\hskip 1em plus 0.5em minus 0.4em\relax IEEE, 2011, pp.
  337--340.

\bibitem{ramos2009svm}
R.~Ramos-Lara, M.~L{\'o}pez-Garc{\'\i}a, E.~Cant{\'o}-Navarro, and
  L.~Puente-Rodriguez, ``Svm speaker verification system based on a low-cost
  fpga,'' in \emph{2009 International Conference on Field Programmable Logic
  and Applications}.\hskip 1em plus 0.5em minus 0.4em\relax IEEE, 2009, pp.
  582--586.

\bibitem{mandal2014implementation}
B.~Mandal, M.~P. Sarma, K.~K. Sarma, and N.~Mastorakis, ``Implementation of
  systolic array based svm classifier using multiplierless kernel,'' in
  \emph{2014 International Conference on Signal Processing and Integrated
  Networks (SPIN)}, 2014, pp. 35--39.

\bibitem{xue2020real}
Z.~Xue, J.~Wei, and W.~Guo, ``A real-time naive bayes classifier accelerator on
  fpga,'' \emph{IEEE Access}, vol.~8, pp. 40\,755--40\,766, 2020.

\bibitem{elhoushi2021deepshift}
M.~Elhoushi, Z.~Chen, F.~Shafiq, Y.~H. Tian, and J.~Y. Li, ``Deepshift: Towards
  multiplication-less neural networks,'' in \emph{Proceedings of the IEEE/CVF
  Conference on Computer Vision and Pattern Recognition}, 2021, pp. 2359--2368.

\bibitem{lin2016fixed}
D.~Lin, S.~Talathi, and S.~Annapureddy, ``Fixed point quantization of deep
  convolutional networks,'' in \emph{International conference on machine
  learning}.\hskip 1em plus 0.5em minus 0.4em\relax PMLR, 2016, pp. 2849--2858.

\bibitem{zhou2018adaptive}
Y.~Zhou, S.-M. Moosavi-Dezfooli, N.-M. Cheung, and P.~Frossard, ``Adaptive
  quantization for deep neural network,'' in \emph{Thirty-Second AAAI
  Conference on Artificial Intelligence}, 2018.

\bibitem{hagiescu2019bfloat}
A.~Hagiescu, M.~Langhammer, B.~Pasca, P.~Colangelo, J.~Thong, and N.~Ilkhani,
  ``Bfloat mlp training accelerator for fpgas,'' in \emph{2019 International
  Conference on ReConFigurable Computing and FPGAs (ReConFig)}.\hskip 1em plus
  0.5em minus 0.4em\relax IEEE, 2019, pp. 1--5.

\bibitem{cristianini2000introduction}
N.~Cristianini, J.~Shawe-Taylor \emph{et~al.}, \emph{An introduction to support
  vector machines and other kernel-based learning methods}.\hskip 1em plus
  0.5em minus 0.4em\relax Cambridge university press, 2000.

\bibitem{gu2012theory}
M.~Gu, \emph{Theory, Synthesis and Implementation of Current-mode CMOS
  Piecewise-linear Circuits Using Margin Propagation}.\hskip 1em plus 0.5em
  minus 0.4em\relax Michigan State University, Electrical Engineering, 2012.

\bibitem{sailor2017unsupervised}
H.~B. Sailor, D.~M. Agrawal, and H.~A. Patil, ``Unsupervised filterbank
  learning using convolutional restricted boltzmann machine for environmental
  sound classification.'' in \emph{InterSpeech}, vol.~8, 2017, p.~9.

\bibitem{xu2018fpga}
Y.~Xu, C.~S. Thakur, R.~K. Singh, T.~J. Hamilton, R.~M. Wang, and A.~van
  Schaik, ``A fpga implementation of the car-fac cochlear model,''
  \emph{Frontiers in neuroscience}, vol.~12, p. 198, 2018.

\bibitem{lu2000design}
W.-S. Lu and A.~Antoniou, ``Design of digital filters and filter banks by
  optimization: A state of the art review,'' in \emph{2000 10th European signal
  processing conference}.\hskip 1em plus 0.5em minus 0.4em\relax IEEE, 2000,
  pp. 1--4.

\bibitem{soares2015approximate}
L.~B. Soares, S.~Bampi, and E.~Costa, ``Approximate adder synthesis for
  area-and energy-efficient fir filters in cmos vlsi,'' in \emph{2015 IEEE 13th
  International New Circuits and Systems Conference (NEWCAS)}.\hskip 1em plus
  0.5em minus 0.4em\relax IEEE, 2015, pp. 1--4.

\bibitem{greenwood1990cochlear}
D.~D. Greenwood, ``A cochlear frequency-position function for several
  species—29 years later,'' \emph{The Journal of the Acoustical Society of
  America}, vol.~87, no.~6, pp. 2592--2605, 1990.

\bibitem{mahmoodi2011fpga}
D.~Mahmoodi, A.~Soleimani, H.~Khosravi, M.~Taghizadeh \emph{et~al.}, ``Fpga
  simulation of linear and nonlinear support vector machine,'' \emph{Journal of
  Software Engineering and Applications}, vol.~4, no.~05, p. 320, 2011.

\bibitem{cutajar2013hardware}
M.~Cutajar, E.~Gatt, I.~Grech, O.~Casha, and J.~Micallef, ``Hardware-based
  support vector machine for phoneme classification,'' in \emph{Eurocon
  2013}.\hskip 1em plus 0.5em minus 0.4em\relax IEEE, 2013, pp. 1701--1708.

\bibitem{boujelben2018efficient}
O.~Boujelben and M.~Bahoura, ``Efficient fpga-based architecture of an
  automatic wheeze detector using a combination of mfcc and svm algorithms,''
  \emph{Journal of Systems Architecture}, vol.~88, pp. 54--64, 2018.

\bibitem{khosravi2007introducing}
H.~Khosravi and E.~Kabir, ``Introducing a very large dataset of handwritten
  farsi digits and a study on their varieties,'' \emph{Pattern recognition
  letters}, vol.~28, no.~10, pp. 1133--1141, 2007.

\bibitem{garofolo1993darpa}
J.~S. Garofolo, L.~F. Lamel, W.~M. Fisher, J.~G. Fiscus, and D.~S. Pallett,
  ``Darpa timit acoustic-phonetic continous speech corpus cd-rom. nist speech
  disc 1-1.1,'' \emph{NASA STI/Recon technical report n}, vol.~93, p. 27403,
  1993.

\bibitem{chollet2015keras}
F.~Chollet \emph{et~al.}, ``Keras,'' 2016, https://github.com/fchollet/keras.

\bibitem{abadi2016tensorflow}
M.~Abadi, P.~Barham, J.~Chen, Z.~Chen, A.~Davis, J.~Dean, M.~Devin,
  S.~Ghemawat, G.~Irving, M.~Isard \emph{et~al.}, ``Tensorflow: A system for
  large-scale machine learning,'' in \emph{12th $\{$USENIX$\}$ Symposium on
  Operating Systems Design and Implementation ($\{$OSDI$\}$ 16)}, 2016, pp.
  265--283.

\bibitem{shanker1996effect}
M.~Shanker, M.~Y. Hu, and M.~S. Hung, ``Effect of data standardization on
  neural network training,'' \emph{Omega}, vol.~24, no.~4, pp. 385--397, 1996.

\end{thebibliography}

\end{document}